\documentclass[letterpaper]{article} 
\usepackage[preprint]{aaai2027}  
\usepackage[hyphens]{url}  
\usepackage{graphicx} 
\usepackage{natbib}  
\usepackage{caption} 
\usepackage{algorithm}
\usepackage{amsmath}
\usepackage{amssymb}
\usepackage{mathtools}
\usepackage{amsthm}
\usepackage{amssymb,amsfonts}
\usepackage{booktabs}
\usepackage{multirow}
\usepackage{makecell}
\usepackage{multicol}
\usepackage{multirow}
\usepackage{graphicx}
\usepackage{booktabs}
\usepackage{caption}
\usepackage{graphicx}
\usepackage{subcaption}
\usepackage{algorithm}
\usepackage{algpseudocode}
\usepackage{amsmath}  
\usepackage{amssymb}  
\usepackage{xcolor}
\usepackage{amsthm}
\newtheorem{proposition}{Proposition}
\usepackage{tcolorbox}

\usepackage{url}
\usepackage{newfloat}
\usepackage{listings}
\DeclareCaptionStyle{ruled}{labelfont=normalfont,labelsep=colon,strut=off} 
\floatstyle{ruled}
\newfloat{listing}{tb}{lst}{}
\floatname{listing}{Listing}

\usepackage{booktabs}

\title{Tripwire: Triggering Aligned Refusal via Statistically Certified Safety Neurons}
\author{
    Wei Zhao, Zhe Li, Peixin Zhang, Jun Sun 
}
\affiliations{
   Singapore Management University
   \\
    \texttt{\{wzhao,zheli,pxzhang,junsun\}@smu.edu.sg}
}

\begin{document}
\providecommand{\ours}{\textsc{TripWire}}

\maketitle

\begin{abstract}
Neuron- and path-level interventions offer the finest-grained route to defending large language models (LLMs) against jailbreak attacks, yet existing methods fall short of this promise, i.e., they often compromise model utility significantly. Specifically, one line of work suppresses toxic neurons to erase harmful semantics, but since such semantics are distributed across the network, blocking every pathway forces a large intervention footprint. An alternative line of research focus on identify safety neurons using external classifiers. While promising, the existing  approaches suffer from compromising neurons that are important for the model utility as well. Moreover, both approaches remain always on and thus perturb every benign request even when no attack is present. To address these limitations, we present \ours{}, a training-free defense that first identifies safety-specific neurons through per-neuron hypothesis tests under false-discovery-rate control together with a utility-specificity filter. Based on this identification, a trigger-style clamp holds the selected neurons at their harmful-conditional mean activations, injecting an internal harmful-input signal that triggers the refusal behavior learned during alignment. The clamp is then realized by two provably equivalent deployment modes, namely a detector-gated inference-time intervention and an offline bias-patch weight edit. Extensive experiments across four safety-aligned LLMs and four representative attacks demonstrate that \ours{} reduces the average attack success rate to at most 2.0\% while incurring a utility drop of only 0.5\% to 5.3\% on MT-Bench, the smallest among all defenses. Code is available at https://anonymous.4open.science/r/Tripwire-65C4. 
\end{abstract}

\section{Introduction}
\providecommand{\ours}{\textsc{TripWire}}
Large Language Models (LLMs) such as ChatGPT~\cite{gpt4} and Gemini~\cite{team2023gemini} have revolutionized natural language processing, achieving remarkable performance in tasks such as question answering, code completion, and text generation~\cite{brown2020language,chen2021evaluating}. However, as these models are increasingly deployed in real-world applications, their vulnerabilities pose significant security and reliability risks. A particularly severe threat is \emph{jailbreaking}, i.e., the use of adversarial prompts to bypass safety mechanisms and elicit harmful or policy-violating outputs~\cite{jailbreakr1_2025,feng2026lookahead}. Such attacks have become increasingly automated and effective: optimization-based attacks such as GCG~\cite{GCG23}, iterative refinement attacks such as PAIR~\cite{PAIR23}, and generator-based attacks such as AmpleGCG~\cite{amplegcg2024} and Jailbreak-R1~\cite{jailbreakr1_2025} have all demonstrated increasing efficiency and effectiveness in bypassing safety mechanisms.

In response, the research community has developed a variety of defense strategies ranging from knowledge-editing approaches such as LED~\cite{LED2405} and DELMAN~\cite{2025delman} to inference-level defenses that examine or perturb inputs and internal states~\cite{SmoothLLM2310,SafeDecoding2402}. Despite these advances, existing defenses exhibit a persistent safety and utility trade-off: strengthening refusal behavior degrades performance on benign tasks. An ideal defense mechanism would operate at the finest possible granularity: by first identifying the safety-related neurons and then intervening on only those neurons, such a defense should achieve strong protection with the least utility cost~\cite{2025neurostrike,safeneuron2025,tracerouter2026}. However, existing neuron-level methods fall short of this promise in practice for three reasons.

First, existing methods such as TraceRouter~\cite{tracerouter2026} and DELMAN~\cite{2025delman} essentially target \emph{toxic neurons}: they rely on harmful vocabulary to locate the modules that encode harmful information and then zero out or reverse the neuron values to block the propagation of harmful semantics. However, harmful semantics are distributed across the network and blocking them requires suppressing every route they may take. This demands a large intervention footprint and turns every additional suppressed neuron into a potential utility loss. Moreover, evolving attacks can always exploit an uncovered route, which makes it practically impossible to invalidate all pathways. Second, methods such as NeuroStrike~\cite{2025neurostrike} instead search for \emph{safety neurons} that detect harmful requests and trigger refusal. These methods rely on external classifiers to attribute neuron importance. However, such classifier-based attribution cannot distinguish safety neurons from generally important ones. Third, some neurons inevitably serve both safety and general functions. An always-on intervention without taking utility into consideration  on these neurons therefore perturbs every benign request even when no attack is present. These three problems are the causes of the observed utility cost rather than the neuron granularity itself.

To address these challenges, we present \ours{}, a neuron-level defense framework that addresses each problem with a dedicated design. First, a \emph{statistically rigorous identification funnel} screens every neuron in the model with direction filtering, per-neuron Welch $t$-tests under Benjamini--Hochberg false-discovery-rate control, and a utility-specificity filter that removes neurons which also activate strongly on normal tasks. The surviving neurons are then ranked and the top-$N$ candidates are selected. This yields a statistical upper bound on the fraction of mis-selected neurons and disentangles safety-specific neurons from generally important ones without any training. Second, a \emph{trigger-style clamp} holds the selected neurons at their harmful-conditional mean activations, which injects an internal ``the current input is harmful'' signal and triggers the refusal behavior the model already learned during alignment. Because triggering refusal requires only a stable signal rather than interception of every semantic pathway, a small neuron set is sufficient and the intervention does not depend on the specific attack. Third, \emph{two deployment modes} realize the same clamp: an inference mode in which a lightweight detector gates the clamp so benign inputs are processed by the unmodified model, and an offline \emph{bias-patch} weight edit shown to be equivalent to the inference-time clamp, both theoretically and empirically. With this design, a single identification pipeline serves both a detector-gated deployment and a permanent, detector-free, zero-overhead checkpoint.

To validate the effectiveness of \ours{}, we conduct comprehensive experiments on four safety-aligned open-source LLMs (Llama-2-7B, Llama-3.1-8B, Qwen2.5-7B, and Qwen2.5-32B) against four representative jailbreak attacks (GCG, AmpleGCG, AutoDAN, and Jailbreak-R1). We compare against state-of-the-art defense baselines organized  by deployment form: RepE~\cite{Representation2023Zou} and TraceRouter~\cite{tracerouter2026} for the inference-based defense and LED~\cite{LED2405} and DELMAN~\cite{2025delman} for the weight-editing defense. We report the attack success rate (ASR) together with MT-Bench~\cite{zheng2023judging} utility scores judged by GPT-5.4 at neuron budgets of top-1000 and top-2500. Results demonstrate that \ours{} (when targeting top-2500 selected neurons) achieves an average ASR below 2\% across both deployment forms on Llama-2, Llama-3.1, and Qwen2.5, compared to 8.3\%--12.6\% for RepE and 2.2\%--4.9\% for TraceRouter, DELMAN, and LED, while incurring the smallest utility drop among all defenses. The inference-based and edit-based forms yield consistent results, confirming that the identified neuron set rather than the specific intervention mechanism carries the defense.


\section{Preliminary}
\subsection{Large Language Models and Neurons}
\label{sec:pre_llm}
An LLM is a decoder-only Transformer that autoregressively maps a token sequence to a distribution over the next token. Each input token is first embedded into a vector $x \in \mathbb{R}^{d}$ and then processed through $L$ stacked layers, each of which updates the representation as
\begin{equation}
x^{(\ell)} = x^{(\ell-1)} + \mathrm{Attn}^{(\ell)}\!\big(x^{(\ell-1)}\big) + \mathrm{MLP}^{(\ell)}\!\big(x^{(\ell-1)}\big),
\end{equation}
where the attention sublayer $\mathrm{Attn}^{(\ell)}$ exchanges information across token positions and the MLP sublayer $\mathrm{MLP}^{(\ell)}$ transforms each position independently. Since the MLP sublayers hold the majority of the parameters and are widely regarded as the primary store of knowledge, we center our analysis on the MLP sublayers.

Modern LLMs such as Llama and Qwen implement the MLP sublayer in a gated form:
\begin{equation}
\begin{aligned}
\mathrm{MLP}^{(\ell)}(x) &= W_{\mathrm{down}}^{(\ell)}\Big(\sigma\big(W_{\mathrm{gate}}^{(\ell)} x\big) \odot \big(W_{\mathrm{up}}^{(\ell)} x\big)\Big), \\
& W_{\mathrm{gate}}^{(\ell)}, W_{\mathrm{up}}^{(\ell)} \in \mathbb{R}^{d_{\mathrm{ff}} \times d}, \quad W_{\mathrm{down}}^{(\ell)} \in \mathbb{R}^{d \times d_{\mathrm{ff}}},
\end{aligned}
\end{equation}
where $d_{\mathrm{ff}}$ is the intermediate dimension of the MLP sublayer, $\sigma$ is a nonlinearity, and $\odot$ denotes the element-wise product. In this work, we define a \emph{neuron} as a triple $(\ell, \psi, i)$ with $\psi \in \{\mathrm{gate}, \mathrm{up}\}$ and $i \in [d_{\mathrm{ff}}]$, whose activation is the $i$-th component of the corresponding projection:
\begin{equation}
a^{(\ell,\psi)}_i(x) = \big(W_{\psi}^{(\ell)} x\big)_i = \big(w^{(\ell,\psi)}_i\big)^{\top} x ,
\end{equation}
so that each neuron corresponds to one row of $W_\psi^{(\ell)}$, which underlies the intervention and editing operations described in the following sections.

We represent each prompt $p$ on neuron $(\ell,\psi,i)$ by its \emph{last-token} activation $a^{(\ell,\psi)}_i(p) \triangleq a^{(\ell,\psi)}_i(x_{t_p})$, where $t_p$ is the last-token position. We choose the last-token position because the causal attention of the decoder-only architecture aggregates the semantic information of the entire prompt. We collect such activations over three prompt sets: a harmful set $\mathcal{D}_h$ from HarmBench~\cite{GCG23}, a benign set $\mathcal{D}_s$ from Alpaca~\cite{alpaca}, and a utility set $\mathcal{D}_u$ from MMLU~\cite{hendrycks2021mmlu}, each containing $200$ prompts.

\begin{figure*}[ht]
    \centering
    \includegraphics[width=0.9\textwidth]{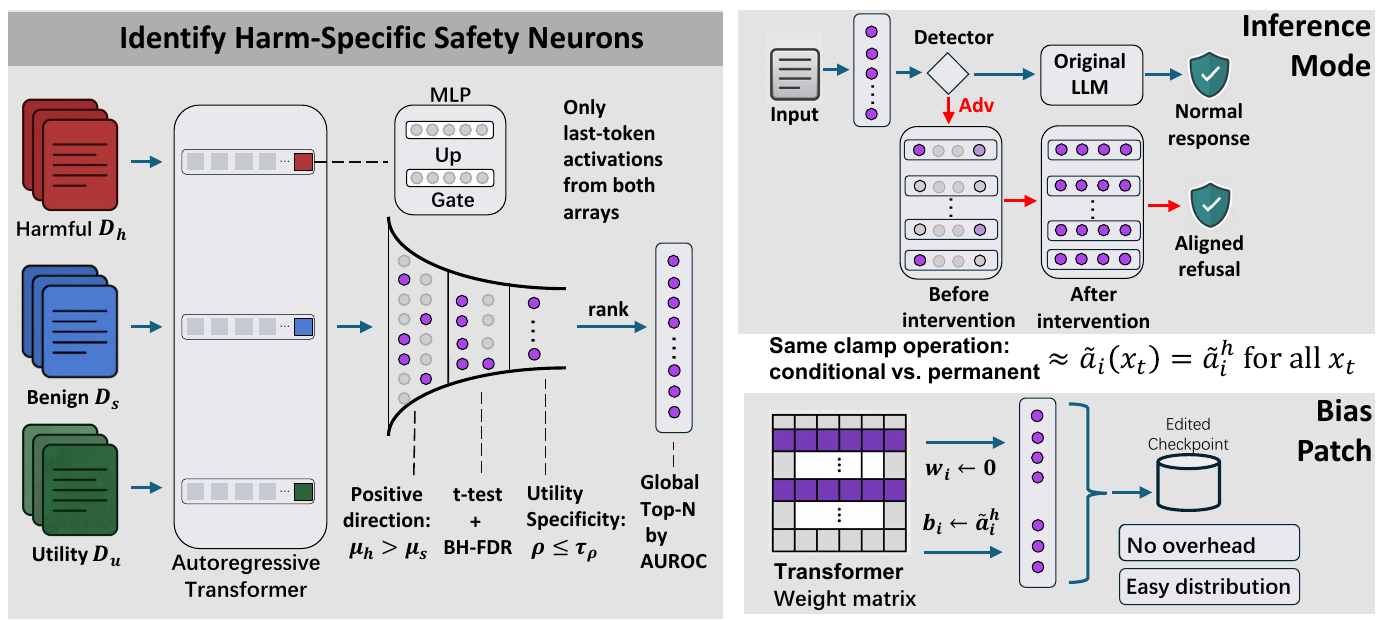}
    \caption{Overview of \ours{}. A statistically rigorous funnel identifies harm-specific safety neurons, a trigger-style clamp injects the harmfulness signal to trigger aligned refusal, and two deployment modes realize the defense with near-zero utility cost.}
    \label{fig:overview}
\end{figure*}

\subsection{Identifying Safety Neurons}
\label{sec:pre_ident}
Previous studies have shown that the safety behavior of an aligned LLM is governed by a remarkably small fraction of neurons: pruning less than $0.6\%$ of the neurons in a layer suffices to break safety alignment~\cite{2025neurostrike,2025neurons}. We conjecture that these \emph{safety neurons} serve two complementary roles: a \emph{detection} role, in which a neuron activates selectively on harmful input and flags the current request as unsafe, and a \emph{refusal} role, in which the sustained activation of these neurons triggers the downstream computation that produces an explicit refusal. This conjecture is consistent with the observation above: once these neurons are set to zero, both roles are lost, and the model no longer refuses harmful requests.

To identify safety neurons, existing work requires training an external classifier on paired harmful and benign prompts. The representative approach fits a per-layer logistic-regression probe that predicts the safety label from the layer's activations~\cite{2025neurostrike},
\begin{equation}
\hat{y}(\mathbf{x}) = \sigma\!\Big(\textstyle\sum_{i} w_i\, a_i^{(\ell)}(\mathbf{x}) + b\Big),
\end{equation}
and selects the neurons whose weight magnitude $|w_i|$ passes a z-score threshold. This identification, however, suffers from two drawbacks. First, the probe weights are not identifiable in the $p \gg n$ regime. A high-weight neuron may therefore be merely correlated with the safety judgment instead of being specific to it, and the fraction of falsely selected neurons is left uncontrolled. Second, the screening does not separate \emph{safety-specific} neurons from \emph{generally important} ones, so intervening on a mislabeled neuron costs utility for no defensive gain. These drawbacks directly motivate the identification funnel of \ours{}.

\section{Method}
\label{sec:method}
In this section, we describe the three components of \ours{} as illustrated in Figure~\ref{fig:overview}.

\subsection{Statistically Rigorous Identification Funnel}
\label{sec:method_funnel}
We treat every neuron in the model as a candidate safety neuron and screen them jointly across all layers. For each neuron independently, we compute class-discrimination statistics over its activation samples $\{a_j^h\}, \{a_j^s\}, \{a_j^u\}$ on $\mathcal{D}_h, \mathcal{D}_s, \mathcal{D}_u$, with means $\mu_h, \mu_s, \mu_u$ and standard deviations $s_h, s_s$. We select a neuron as a safety neuron only if it passes three successive filters: (i) direction, (ii) significance, and (iii) specificity. The neurons that pass all three filters are then ranked, and the top ones are selected under a specified budget $N$.

\textbf{(i) Direction filtering.} We retain only neurons \emph{positively} correlated with harmful content: $\mu_h > \mu_s$.

\textbf{(ii) Significance filtering (Welch $t$-test + BH-FDR).} For each neuron we perform a heteroscedastic two-sample Welch $t$-test,
\begin{equation}
t = \frac{\mu_h - \mu_s}{\sqrt{s_h^2/n_h + s_s^2/n_s}},
\end{equation}
and correct the resulting $p$-values across \emph{all} neurons of \emph{all} layers with the Benjamini--Hochberg procedure: sorting $p$-values in ascending order $p_{(1)} \le \cdots \le p_{(m)}$ and rejecting the first $k^\ast = \max\{k : p_{(k)} \le \frac{k}{m}\alpha\}$ hypotheses at $\alpha = 0.05$. Under the standard assumptions of the BH procedure, this controls
the expected fraction of irrelevant neurons admitted into the set
selected at level $\alpha$.
This is the first layer of utility protection, since every mis-selected neuron contributes nothing to defense and only damages normal capabilities.

\textbf{(iii) Utility-specificity filtering.} To suppress over-refusal (i.e., wrongly rejecting safe request~\cite{cui2024or}) and protect general capabilities, we measure where a neuron's utility-task activation lies on the benign-to-harmful axis:
\begin{equation}
\rho = \frac{\mu_u - \mu_s}{\mu_h - \mu_s}.
\end{equation}
$\rho \approx 0$ indicates that the neuron behaves on utility tasks as it does on benign inputs, whereas $\rho \approx 1$ indicates that it also activates strongly on normal utility tasks, i.e., a \emph{generally important} neuron rather than a safety-specific one. We exclude neurons with $\rho > \tau_\rho$ (default $\tau_\rho = 0.5$).

\textbf{Ranking and selection budget.} We rank the remaining candidates by AUROC, computed via ranks on the pooled sample $\{a_j^h\} \cup \{a_j^s\}$ (average ranks for ties), which equals the normalized Mann--Whitney $U$ statistic:
\begin{equation}
\begin{aligned}
\mathrm{AUROC} &= \frac{1}{n_h n_s}\left(\sum_{j=1}^{n_h} \mathrm{rank}\big(a_j^h\big) - \frac{n_h(n_h+1)}{2}\right) \\
&= \Pr\big(a^h > a^s\big),
\end{aligned}
\end{equation}
i.e., the probability that a randomly drawn harmful activation exceeds a randomly drawn benign one. This univariate effect size provides an independent, interpretable, and scale-invariant measure of discriminative strength for each neuron.

Because the harmful and benign distributions differ statistically on almost every neuron, an absolute AUROC threshold cannot cleanly delimit the distribution tail. We therefore impose a fixed selection budget and take the global top-$N$ by AUROC over the candidate set $\mathcal{C}$ that passes (i)--(iii):
\begin{equation}
\mathcal{S} = \operatorname*{arg\,top\text{-}N}_{(\ell,\psi,i)\, \in\, \mathcal{C}} \mathrm{AUROC}^{(\ell,\psi,i)},
\end{equation}
yielding per-layer index sets $\mathcal{S}_\ell^{\psi} \subseteq [d_{\mathrm{ff}}]$.

\textbf{Comparison with probe-based selection.} Probe-based attribution~\cite{2025neurostrike} fits a per-layer logistic classifier and selects neurons by weight z-scores. These weights are not identifiable in the $p \gg n$ regime. A high-weight neuron may therefore be merely correlated with harmfulness instead of being specific to it, and no false-discovery guarantee holds. Our funnel replaces this unidentifiable joint attribution with per-neuron univariate statistics under explicit FDR and specificity control. In the following section, we compare the safety and utility performance of the neurons selected by our funnel against those selected by this probe-based method.

\subsection{Trigger-Style Clamp Intervention}
\label{sec:method_clamp}
Given the selected set $\mathcal{S}$, we compute for each safety neuron its harmful-conditional mean activation over the harmful set $\mathcal{D}_h$:
\begin{equation}
\bar{a}^{h,(\ell,\psi)}_i = \frac{1}{|\mathcal{D}_h|}\sum_{p \in \mathcal{D}_h} a^{(\ell,\psi)}_i(p).
\end{equation}
The intervention clamps each selected neuron to this constant at all token positions $t$ throughout generation:
\begin{equation}
\tilde{a}^{(\ell,\psi)}_i(x_t) =
\begin{cases}
\bar{a}^{h,(\ell,\psi)}_i, & i \in \mathcal{S}_\ell^\psi,\\[2pt]
a^{(\ell,\psi)}_i(x_t), & \text{otherwise}.
\end{cases}
\label{eq:clamp}
\end{equation}

Erasure-style interventions~\cite{tracerouter2026,2025neurostrike} attempt to block the propagation of harmful semantics. Since these semantics are distributed across the network, blocking them requires suppressing every route they may take. Newly emerging attacks may target different activation paths that are hard to cover in advance. Our clamp does the opposite: it holds the ``harmfulness detector'' neurons at their harmful-conditional means throughout generation. The clamp overwrites the activations directly and thus guarantees that no adversarial prompt can suppress this signal. In this way, the model continuously receives an internal signal that the current input is a harmful request, which triggers the refusal behavior the model already acquired during alignment. Because triggering refusal requires only a stable signal rather than the interception of every pathway, a small and rigorously selected neuron set is sufficient and the intervention does not depend on the specific attack.

\subsection{Two Deployment Modes}
\label{sec:method_deploy}

\subsubsection{Inference Mode}
\label{sec:method_inference}
Although our funnel separates safety-specific neurons from generally important ones as much as possible, some selected neurons may still carry utility knowledge and an always-on clamp would inevitably perturb benign requests. To avoid this cost, we build a sparse feature from the selected neuron set and train a classifier on the identification prompt sets to decide whether to intervene. The inference mode therefore applies the clamp \emph{only} to requests judged malicious, so that benign inputs are processed by the unmodified model.

\textbf{Sparse feature construction.} For an input prompt $p$, we read only the selected safety neurons' last-token activations and concatenate them in a fixed layer order:
\begin{equation}
\phi(p) = \bigoplus_{\ell \in \mathcal{L},\, \psi} \Big[a^{(\ell,\psi)}_i(p)\Big]_{i \in \mathcal{S}_\ell^\psi} \in \mathbb{R}^{D}, \qquad D \ll2 L \cdot d.
\end{equation}

\textbf{Detector training.} To classify malicious requests, we employ logistic regression as a linear classifier over the sparse features:
\begin{equation}
P(\mathrm{adv} \mid p) = \sigma\big(\theta^\top \phi(p) + \theta_0\big),
\end{equation}
where $\sigma$ is the sigmoid function. The classifier is trained on the same benign set $\mathcal{D}_s$ and harmful set $\mathcal{D}_h$ used for identification.

\textbf{Detect-then-intervene.} For each request $p$, a single forward pass extracts $\phi(p)$. If the trained detector judges the request as benign, the model generates normally; if it judges the request as harmful, the clamp of Eq.~\eqref{eq:clamp} is applied via forward hooks throughout generation. Detection and intervention share the same neuron set and analysis data, so the two stages are naturally self-consistent.

\begin{figure*}[t]
\centering
\includegraphics[width=0.95\textwidth]{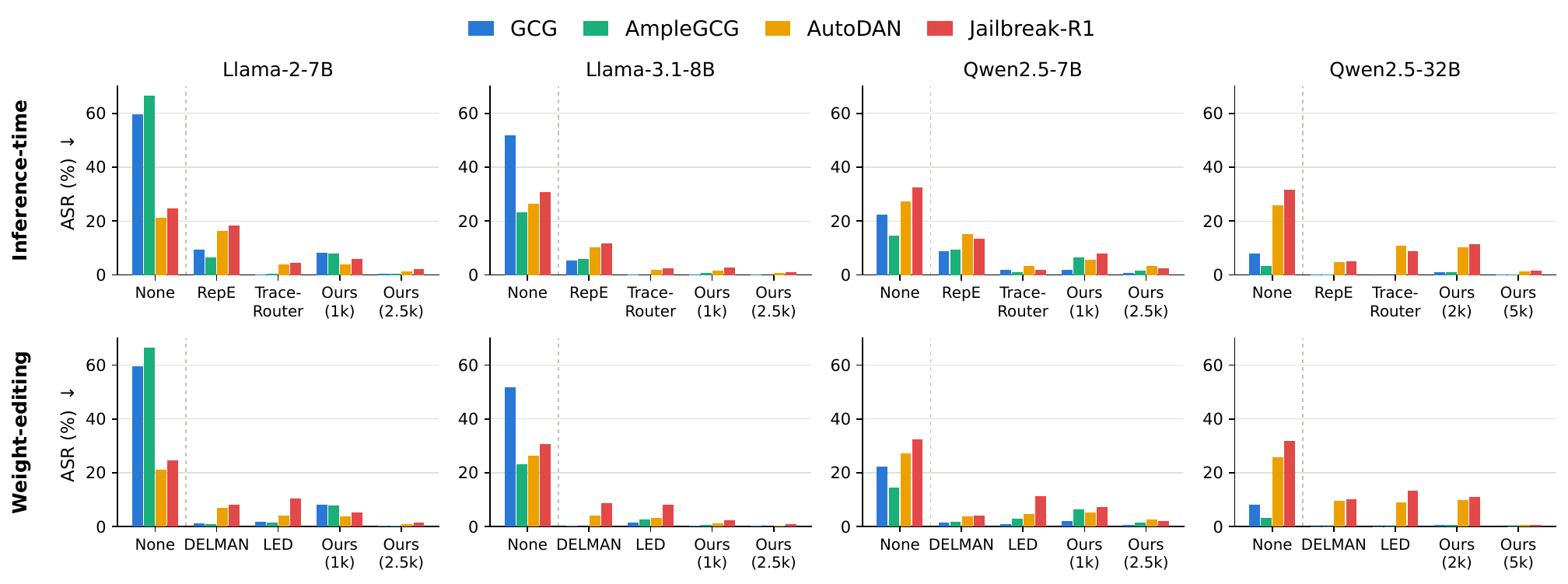}
\caption{ASR (\%, $\downarrow$) under four jailbreak attacks for all defenses across four models. Top row: inference-time defenses; bottom row: weight-editing defenses.}
\label{fig:main_results}
\end{figure*}

\subsubsection{Offline Mode}
\label{sec:method_edit}
Alternatively, if the goal is to deliver a hardened checkpoint that is always-on and needs no detector at inference time, we write the equivalent defense directly into the weights. The always-on clamp of Eq.~\eqref{eq:clamp} admits an exact weight-space realization that we call the \emph{bias-patch}. For each selected neuron $(\ell, \psi, i) \in \mathcal{S}$, with weight row $w_i$ and bias $b_i$ (a $b=0$ bias parameter is injected when the architecture has none), we zero the weight row and inject the harmful-conditional mean as the bias:
\begin{equation}
w_i \leftarrow \mathbf{0}, \qquad b_i \leftarrow \bar{a}^h_i
\quad\Longrightarrow\quad \tilde{a}_i(x) \equiv \bar{a}^h_i.
\label{eq:biaspatch}
\end{equation}

\begin{proposition}[Equivalence of bias-patch and always-on  inference-time clamp]
For every input $x$ and token position $t$, the bias-patched model computes exactly the activations of Eq.~\eqref{eq:clamp}: zeroing $w_i$ makes $a_i(x) = w_i^\top x + b_i \equiv b_i = \bar{a}^h_i$ for $i \in \mathcal{S}_\ell^\psi$, while all other neurons are untouched. Hence the bias-patched checkpoint and the inference-time clamped model are mathematically identical maps, differing only in that the former is permanent and detector-free.
\end{proposition}

\noindent The bias-patch is thus the exact bridge between the two deployment routes, so any defense property established for the inference-time clamp remains equivalent on the edited checkpoint. The choice between the two modes is then purely operational: the inference mode preserves benign behavior exactly at the cost of a per-request gating decision, whereas the bias-patch mode requires no detector but applies the clamp unconditionally. Because the clamped set is small and harm-specific by construction, the always-on cost of the bias-patch remains a low trade-off on utility benchmarks, as quantified in experiment section.


\section{Experiment}
\label{sec:experiments}
\subsection{Experimental Setup}
\textbf{Models.} We conduct experiments on four safety-aligned open-source LLMs: Llama-2-7B~\cite{touvron2023llama}, Llama-3.1-8B~\cite{dubey2024llama}, Qwen2.5-7B, and Qwen2.5-32B~\cite{bai2023qwena}. All experiments are conducted on H100-80GB servers.

\textbf{Attacks.} We evaluate against four representative jailbreak attacks covering diverse attack paradigms: GCG~\cite{GCG23} (white-box gradient optimization), AmpleGCG~\cite{amplegcg2024} (generator-based), AutoDAN~\cite{PAIR23} (black-box iterative refinement), and Jailbreak-R1~\cite{jailbreakr1_2025} (reinforcement-learning driven). For each target model, we follow each attack's original implementation and generate adversarial prompts from the $520$ harmful prompts of AdvBench~\cite{GCG2023Zou} within the setup budget. Detailed attack configurations are provided in the Appendix.

\textbf{Defense Baselines.} We compare \ours{} against state-of-the-art defense methods in both deployment forms. For \emph{inference-time} defenses, we compare against RepE~\cite{CircuitBreaker24} and TraceRouter~\cite{tracerouter2026}, both of which intervene on internal computation at inference time. For \emph{weight-editing} defenses, we compare against LED~\cite{LED2405} and DELMAN~\cite{2025delman}, both of which permanently modify model parameters. We adopt the default parameters from each method. To the best of our knowledge, \ours{} is the only method that covers both deployment forms with a single identification pipeline. This enables a controlled comparison where the neuron set is held fixed and only the deployment form varies. We report \ours{} at two neuron budgets, top-1000 and top-2500. For Qwen2.5-32B, we scale the budgets to top-2000 and top-5000 to match its larger neuron count, i.e., $2500/1{,}060{,}864=0.23\%$ for Qwen2.5-7B and $5000/3{,}538{,}944=0.14\%$ for Qwen2.5-32B.

\textbf{Metrics.} Following standard practice~\cite{GCG23,CircuitBreaker24}, we quantify defense effectiveness through Attack Success Rate (ASR), where a prompt is considered successfully jailbroken if the model generates harmful content that violates safety guidelines; we employ GPT-5.4 as the safety evaluator over (prompt, response) pairs. We quantify utility with MT-Bench~\cite{zheng2023judging} scores, also judged by GPT-5.4. In addition, we report the MT-Bench over-refusal rate by matching refusal keywords such as ``cannot'' over benign responses. Prompt templates for the judge are provided in the Appendix.

\begin{table*}[t]
\centering
\small
\setlength{\tabcolsep}{3.5pt}
\renewcommand{\arraystretch}{1.05}
\caption{Defense--utility trade-off: average ASR (\%, $\downarrow$) over the four attacks, MT-Bench score ($\uparrow$), MT-Bench over-refusal rate (\%, $\downarrow$), and inference-time overhead ($\downarrow$) relative to the undefended model. \textbf{Bold} marks the best result among defenses within each deployment form.}
\label{tab:utility}
\scalebox{0.85}{
\begin{tabular}{ll|ccc|ccc|ccc|ccc|c}
\toprule
& & \multicolumn{3}{c|}{\textbf{Llama-2-7B}} & \multicolumn{3}{c|}{\textbf{Llama-3.1-8B}} & \multicolumn{3}{c|}{\textbf{Qwen2.5-7B}} & \multicolumn{3}{c|}{\textbf{Qwen2.5-32B}} & \\
\textbf{Form} & \textbf{Defense} & ASR & MT-B & Rej. & ASR & MT-B & Rej. & ASR & MT-B & Rej. & ASR & MT-B & Rej. & \textbf{Overhead} \\
\midrule
--- & No defense & 43.0\% & 4.94 & 0.0\% & 33.1\% & 6.60 & 0.0\% & 24.1\% & 7.40 & 0.0\% & 17.2\% & 8.23 & 0.0\% & $1.00\times$ \\
\midrule
\multirow{4}{*}{Inference}
 & RepE & 12.6\% & 3.76 & 15.0\% & 8.3\% & 3.63 & 65.0\% & 11.8\% & 4.52 & 31.3\% & 2.6\% & 6.18 & 33.8\% &$1.27\times$ \\
 & TraceRouter & 2.2\% & 4.19 & 2.5\% & 1.2\% & 5.59 & 5.6\% & \textbf{1.9\%} & 5.92 & 2.5\% & 4.9\% & 7.35 & 7.5\% & $1.16\times$ \\
 & \ours{} (Small) & 6.5\% & \textbf{4.76} & \textbf{1.3\%} & 1.3\% & \textbf{6.56} & \textbf{2.5\%} & 5.6\% & \textbf{7.40} & \textbf{0.0\%} & 5.9\% & \textbf{8.20} & \textbf{2.5\%} & $1.13\times$ \\
 & \ours{} (Large) & \textbf{1.1\%} & 4.68 & 2.5\% & \textbf{0.5\%} & 6.48 & 3.1\% & 2.0\% & 7.36 & 3.8\% & \textbf{0.8\%} & 8.14 & 3.8\% & $1.15\times$  \\
\midrule
\multirow{4}{*}{Edit}
 & DELMAN & 4.3\% & 4.53 & 8.8\% & 3.3\% & \textbf{5.78} & 15.0\% & 2.8\% & 6.16 & \textbf{5.0\%} & 5.0\% & 7.45 & 13.1\% & $1.00\times$ \\
 & LED & 4.4\% & 4.28 & 8.8\% & 3.8\% & 5.26 & 22.5\% & 4.9\% & 5.88 & 11.3\% & 5.6\% & 7.68 & 18.8\% & $1.00\times$ \\
 & \ours{} (Small) & 6.3\% & \textbf{4.73} & \textbf{2.0\%} & 1.1\% & 5.10 & \textbf{8.8\%} & 5.3\% & \textbf{6.96} & 8.0\% & 5.6\% & \textbf{7.88} & \textbf{8.8\%} & $1.00\times$ \\
 & \ours{} (Large) & \textbf{0.8\%} & 4.31 & 15.0\% & \textbf{0.3\%} & 4.84 & 31.3\% & \textbf{1.7\%} & 5.86 & 12.5\% & \textbf{0.3\%} & 7.45 & 17.5\% & $1.00\times$ \\
\bottomrule
\end{tabular}
}
\end{table*}

\subsection{Defense Performance Analysis}

Defense results are summarized in Figure~\ref{fig:main_results} and Table~\ref{tab:utility}. Figure~\ref{fig:main_results} reports ASR under the four attacks for all defenses across the four models, and Table~\ref{tab:utility} summarizes the defense--utility trade-off through average ASR, MT-Bench score, over-refusal rate, and inference-time overhead. We measure the time overhead over MT-Bench generation, as refusal responses are inherently shorter and offer an unfair comparison.

As shown in Figure~\ref{fig:main_results}, for the undefended models, the ASR of GCG and AmpleGCG varies substantially across models and drops sharply on Qwen2.5-32B, as larger and more recently aligned models present a harder optimization landscape for gradient-based suffix search. In contrast, AutoDAN and Jailbreak-R1 maintain similar ASR across all four models. Among the defenses, all methods except RepE reduce ASR significantly; RepE yields the weakest defense, as its steering direction extracted from mean representation differences introduces excessive noise. TraceRouter and \ours{} at the top-2500 budget suppress ASR to nearly zero across all four models (\ours{} in both the inference-based and edit-based forms). Across different attack methods, most defenses effectively defend against GCG and AmpleGCG. However, the human-readable prompts from AutoDAN and Jailbreak-R1 are harder to defend. Jailbreak-R1 in particular evades many defenses due to its highly diverse attack strategies.

As shown in Table~\ref{tab:utility}, in the inference-based form, \ours{} at the top-2500 budget reduces the average ASR from 43.0\% to 1.1\% on Llama-2, from 33.1\% to 0.5\% on Llama-3.1, and from 24.1\% to 2.0\% on Qwen2.5, with a utility drop of only 5.3\%, 4.7\%, and 0.5\% respectively. TraceRouter reaches a comparable ASR but degrades utility by up to 20.0\%. We attribute this cost to its erasure-style intervention, which suppresses neurons along harmful pathways and inevitably perturbs benign computation that shares those pathways. RepE remains at 8.3\%--12.6\% average ASR and degrades utility by up to 45.0\%. Notably, the inference-based top-1000 configuration of \ours{} keeps the utility drop within 3.6\% and over-refuses at most 2.5\% of benign queries, as the detector leaves the benign path untouched; the ASR at this budget is slightly higher (1.3\%--6.5\%) due to the smaller neuron set. In the edit-based form, \ours{} at top-2500 attains the best average ASR on every model (0.8\%, 0.3\%, and 1.7\%), whereas DELMAN and LED remain at 2.8\%--4.9\%. At top-1000, \ours{} achieves the highest MT-Bench score and the lowest over-refusal rate on Llama-2 and Qwen2.5, though the ASR is moderately higher at 6.3\% and 5.3\% due to the smaller neuron budget. Regarding the time overhead, \ours{} incurs at most $1.15\times$ generation time and remains cheaper than TraceRouter ($1.16\times$) and RepE ($1.27\times$).

Overall, \ours{} at top-2500 achieves an average ASR below 2\% across both deployment forms on Llama-2, Llama-3.1, and Qwen2.5, compared to 8.3\%--12.6\% for RepE and 2.2\%--4.9\% for TraceRouter, DELMAN, and LED, while incurring the smallest utility drop among all defenses.


\subsection{Ablation Study}
To isolate the contribution of the identification funnel, we compare it against probe-based selection~\cite{2025neurostrike} under always-on inference-time clamping, so that the detector is removed and only the selected neuron set differs. We evaluate on Llama-2, Llama-3.1, and Qwen2.5, measuring safety with average ASR and utility with MT-Bench score and over-refusal rate.

Table~\ref{tab:ablation} summarizes the ablation results. We first note that the inference without detector and edit forms of \ours{} yield nearly identical ASR and MT-Bench scores at both budgets, which empirically confirms the mathematical equivalence between the bias-patch and the inference-time clamp established in the preliminary section. The table shows that probe-based selection requires $3$--$5\times$ more neurons than our funnel to reach a comparable ASR. Nevertheless, it degrades utility by up to 52.3\% and triggers far more over-refusal. This is expected, as the probe cannot separate safety-specific neurons from generally important ones. The additional neurons it selects carry utility knowledge and clamping them degrades benign performance. In contrast, our funnel reaches a similar or lower ASR with far fewer neurons and a much milder utility cost. This confirms that the FDR and specificity control in the funnel translates directly into a better safety--utility trade-off. Although the identification step for \ours{} takes 79--92 seconds per model, this is a one-time offline analysis and does not affect inference latency.

\begin{table}[t]
\centering
\small
\setlength{\tabcolsep}{3pt}
\renewcommand{\arraystretch}{1.05}
\caption{Identification ablation under always-on inference-time clamping (no detector gate). Probe-based selection is denoted by its z-score threshold $z$.}
\label{tab:ablation}
\begin{tabular}{l|rr|ccc}
\toprule
\textbf{Variant} & \textbf{Neurons} & \textbf{Extr.} & \textbf{ASR} & \textbf{MT-Bench} & \textbf{Rej.} \\
\midrule
Llama2-7B &      0 & --- & 43.0 & 4.94          & 0.0  \\
 Probe ($z{=}3.0$)  & 3{,}589 & 31s &  3.3 & 4.23          & 8.8  \\
 Probe ($z{=}2.5$)  & 7{,}568 & 32s &  2.9 & 3.82          & 18.8 \\
 \ours{} (top-1000) & 1{,}001 & 79s &  6.3 & \textbf{4.70} & \textbf{2.0}  \\
 \ours{} (top-2500) & 2{,}503 & 80s & \textbf{0.8} & 4.31 & 15.0 \\
\midrule
Llama3.1-8B       &       0 & --- & 33.1 & 6.60          & 0.0  \\
 Probe ($z{=}3.0$)  & 10{,}811 & 43s &  0.1 & 4.08          & 50.0 \\
 Probe ($z{=}2.5$)  & 17{,}234 & 44s &  0.1 & 3.15          & 22.5 \\
 \ours{} (top-1000) &  1{,}000 & 88s &  1.1 & \textbf{5.22} & \textbf{8.8}  \\
 \ours{} (top-2500) &  2{,}500 & 88s & \textbf{0.3} & 4.85 & 31.3 \\
\midrule
Qwen2.5-7B         &       0 & --- & 24.1 & 7.40          & 0.0  \\
Probe ($z{=}3.0$)  &  9{,}664 & 38s &  2.6 & 5.50          & 12.5 \\
 Probe ($z{=}2.5$)  & 15{,}808 & 38s &  0.2 & 3.88          & 62.5 \\
 \ours{} (top-1000) &  1{,}000 & 91s &  5.3 & \textbf{7.02} & \textbf{8.0}  \\
 \ours{} (top-2500) &  2{,}502 & 92s & \textbf{1.7} & 5.87  & 12.5 \\
\bottomrule
\end{tabular}
\end{table}

\begin{figure}[t]
\centering
\includegraphics[width=0.5\textwidth]{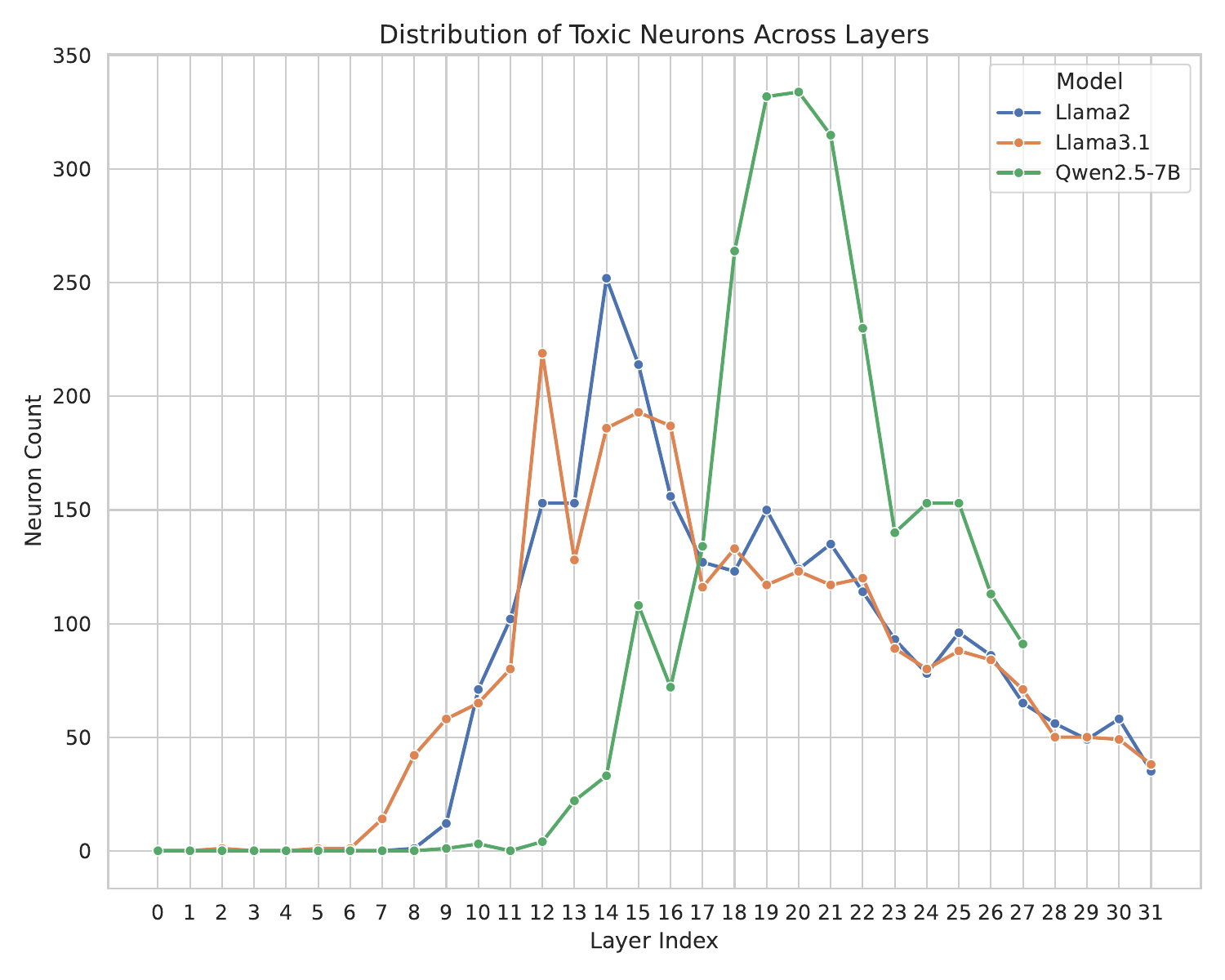}
\caption{Distribution of Toxic Neurons Across Layers}
\label{fig:neuron_dist}
\end{figure}
\subsection{Safety Neuron Distribution Analysis}
To further examine the identified safety neurons, we analyze their distribution across layers for each model. Figure~\ref{fig:neuron_dist} shows the number of selected safety neurons per layer under the top-2500 budget.

Across all three models, safety neurons are concentrated in the middle-to-upper layers and are nearly absent in the early layers. For Llama-2 and Llama-3.1, the neuron count rises sharply around layers 8--14, peaks in the middle layers (layers 12--16), and gradually decreases toward the final layers. For Qwen2.5-7B, the distribution is shifted later, with the majority of safety neurons concentrated in layers 15--21 and a sharp onset around layer 13. This pattern is consistent with the established view that early layers primarily encode low-level syntactic features, while middle and upper layers encode higher-level semantic and behavioral representations. The concentration of safety neurons in these layers suggests that harm detection and refusal triggering are mediated by mid-to-high-level representations rather than surface-level token features.

\section{Related Work}
\label{sec:related}
\subsection{Adversarial Attacks on LLMs}
Jailbreak attacks craft adversarial prompts that circumvent the safety alignment of LLMs, compelling them to produce restricted or harmful outputs. Optimization-based attacks~\cite{GCG23} leverage gradient signals to search for adversarial suffixes that maximize the probability of harmful responses. Generator-based attacks~\cite{amplegcg2024,jailbreakr1_2025} instead fine-tune a model on successful jailbreaks, enabling diverse adversarial prompts to be produced in a single forward pass without further optimization. Iterative refinement attacks~\cite{PAIR23,TAP23,AutoDAN24} employ an attacker LLM to propose, test, and refine human-readable jailbreaks through a feedback-driven loop within a few queries. Besides these methods that perturb the input prompt, another line of work directly targets the internal components of the model, such as representation steering~\cite{2024understanding} and neuron-level manipulation~\cite{2025neurostrike}. Among these component-level methods, NeuroStrike~\cite{2025neurostrike} is closest to our setting: it identifies sparse safety neurons with per-layer logistic-regression probes and prunes them to disable refusal behavior, demonstrating transferability across model variants. From a defense perspective, NeuroStrike adversarially corroborates our core premise that refusal is mediated by a sparse and identifiable set of safety neurons.

\subsection{Defenses for LLMs}
We organize existing defenses by the granularity of their intervention, ranging from the whole model down to individual neurons. At the coarsest granularity, alignment training such as RLHF~\cite{ouyang2022training} and adversarial training~\cite{CAT24,AdversarialTuning2406} instills refusal behavior by shifting the entire response distribution, and must be repeated for every emerging threat. Inference-time defenses instead examine or mutate the input prompt~\cite{Perplexity2308,SmoothLLM2310} or steer the decoding distribution~\cite{SafeDecoding2402}, thereby perturbing every request at each forward pass even when the request is benign. Narrowing to sub-model granularity, Circuit Breaker~\cite{CircuitBreaker24} operates at the representation level, training the model to reroute harmful representations toward orthogonal directions and short-circuit the computation that would otherwise produce unsafe content. LED~\cite{LED2405} operates at the layer level, locating safety-critical layers through pruning analysis and editing their parameters toward safe responses. DELMAN~\cite{2025delman} operates at the parameter level, updating a small set of weights with closed-form solutions regularized by KL divergence to preserve utility. At the finest granularity, TraceRouter~\cite{tracerouter2026} works similarly by locating the neurons along all harm-related pathways and then suppressing or reversing their activations. We emphasize that the neurons TraceRouter targets are those responsible for \emph{generating} harmful content, which are fundamentally different from the safety neurons defined in this work, namely those responsible for \emph{detecting} harmful requests and \emph{triggering} refusal. In addition, unlike these always-on defenses, our inference mode gates the intervention with a detector, so benign requests are processed by the unmodified model and the utility cost is further reduced.

\section{Conclusion}
In this work, we diagnose why existing neuron-level jailbreak defenses fail to deliver the promise of fine-grained protection: erasure-style methods must block harmful semantics distributed across the network, classifier-based attribution cannot separate safety neurons from generally important ones, and always-on interventions perturb benign requests even when no attack is present. These fixable problems, rather than the neuron granularity itself, are the true sources of the observed utility cost.

Based on these insights, we propose \ours{}, a training-free defense that identifies safety-specific neurons through a statistical funnel with false-discovery-rate control and clamps them at their harmful-conditional mean activations to trigger the refusal behavior learned during alignment, realized by two provably equivalent deployment modes. Across four safety-aligned LLMs and four representative attacks, \ours{} reduces the average attack success rate to at most 2.0\% with the smallest utility drop among all defenses, establishing that precise identification combined with trigger-style intervention enables practical neuron-level safety without costly retraining.

\newpage

\bibliography{aaai2027}


\end{document}